\documentclass[conference]{IEEEtran}
\IEEEoverridecommandlockouts
\usepackage{cite}
\usepackage{amsmath,amssymb,amsfonts}
\usepackage{verbatim}
\usepackage{booktabs}
\usepackage{algorithm}
\usepackage{algpseudocode}
\usepackage{multirow}
\usepackage{subcaption}
\usepackage{graphicx}
\usepackage{hyperref}
\usepackage{textcomp}
\usepackage{xspace}
\usepackage{multicol}
\usepackage{epsfig}
\usepackage{xcolor}
\usepackage{algorithmicx}
\usepackage{makecell}
\usepackage{tcolorbox}
\tcbuselibrary{breakable}
\newtcolorbox{textbox}[1]{
    sharp corners,
    boxsep=0mm,
    toptitle=2mm,
    lefttitle=0mm,
    colframe=black!3,
    colback=black!3,
    title={\rule[-2pt]{4.5pt}{10pt}\hspace*{1.5mm}#1},
    fonttitle=\bfseries\itshape\sffamily,
    coltitle=black,
    halign=flush left,
    breakable
}
\def\BibTeX{{\rm B\kern-.05em{\sc i\kern-.025em b}\kern-.08em
    T\kern-.1667em\lower.7ex\hbox{E}\kern-.125emX}}
\begin{document}

\title{Boosting Disfluency Detection with Large Language Model as Disfluency Generator}

\author{
  \IEEEauthorblockN{Zhenrong Cheng}
  \IEEEauthorblockA{\textit{School of Intelligence Science and Technology}\\
  \textit{Peking University}\\
  Beijing, China \\
  chengzhenrong@stu.pku.edu.cn}\\

  \IEEEauthorblockN{Hao Sun}
  \IEEEauthorblockA{\textit{School of Intelligence Science and Technology} \\
  \textit{Peking University} \\
  Beijing, China \\
  sunhao@stu.pku.edu.cn}
\and

  \IEEEauthorblockN{Jiayan Guo}
  \IEEEauthorblockA{\textit{School of Intelligence Science and Technology} \\
  \textit{Peking University} \\
  Beijing, China \\
  guojiayan@pku.edu.cn}\\

  \IEEEauthorblockN{Yan Zhang}
  \IEEEauthorblockA{\textit{School of Intelligence Science and Technology} \\
  \textit{Peking University} \\
  Beijing, China \\
  zhyzhy001@pku.edu.cn}\\

}

\maketitle

\begin{abstract}
Current disfluency detection methods heavily rely on costly and scarce human-annotated data. To tackle this issue, some approaches employ heuristic or statistical features to generate disfluent sentences, partially improving detection performance. However, these sentences often deviate from real-life scenarios, constraining overall model enhancement. In this study, we propose a lightweight data augmentation approach for disfluency detection, utilizing the superior generative and semantic understanding capabilities of large language model (LLM) to generate disfluent sentences as augmentation data. We leverage LLM to generate diverse and more realistic sentences guided by specific prompts, without the need for fine-tuning the LLM. Subsequently, we apply an uncertainty-aware data filtering approach to improve the quality of the generated sentences, utilized in training a small detection model for improved performance. Experiments using enhanced data yielded state-of-the-art results. The results showed that using a small amount of LLM-generated enhanced data can significantly improve performance, thereby further enhancing cost-effectiveness. Our code is available \href{https://github.com/orange-city/ll_gen_dd/tree/main}{here}.
\end{abstract}


\section{Introduction}
Disfluencies, such as repetitions, restarts, repairs, and filled pauses, are common occurrences in human speech. Accurately detecting them is essential for various applications, including Automatic Speech Recognition(ASR) \cite{futami2023streaming}, Dialogue Systems \cite{marie2023disfluency}, and Language Understanding \cite{dinkar2023fillers}. The goal of disfluency detection is to identify the disfluencies in ASR outputs \cite{wang-etal-2020-combining}. As shown in Example \ref{example}, the conventional structure of a disfluency includes the following elements: the reparandum $\mathit{to\  Boston}$, which represents the words intended for removal, the ‘+’ signifying the end of the reparandum, the optional interregnum $\mathit{um,\ I\ mean}$ with filled pauses or discourse markers, and the repair $\mathit{to\ Denver}$ which substitutes \begin{equation}
\normalsize{
    \mathrm{a\ flight\ [\ \overbrace{\mathrm{to\ Boston}}^{reparandum},\ +\  \underbrace{\mathrm{\{ um,\ I\ mean \}}}_{interregnum},\ \overbrace{ \mathrm{to\ Denver}}^{repair}\ ]} }
    \label{example}
\end{equation}the original reparandum. According to the speech typology proposed by Shriberg et al. \cite{shriberg1997prosody}, disfluencies can be classified into three categories: repetition, substitution, and restart.

Approaches to disfluency detection encompass noise channel, translation-based, parser-based, and sequence labeling models, which have demonstrated reasonable performance. However, these methods are limited by the scarcity of disfluency data. Recent research has highlighted the benefits of data augmentation, such as adding simulated disfluent data and employing self-training techniques \cite{wang-etal-2020-combining,jamshid-lou-johnson-2020-improving,wang2020multi,rocholl2021disfluency}. The previous data augmentation methods typically relied on simple heuristic rules such as random repetition, insertion or deletion of n-gram grammar \cite{passali-etal-2022-lard}, or extracting statistical features from sentences to generate disfluency data \cite{yang-etal-2020-planning}. These methods tend to produce simple types of disfluency sentences (a sentence has only one place or type of disfluency), and their ability to generate disfluency types related to semantics is limited, resulting in sentences lacking authenticity.

 Large language models (LLM) are renowned for their robust generative capabilities and semantic understanding \cite{zhao2023survey,sun2023towards,sun2023allies}. Recently, Marie et al. \cite{marie2023disfluency}  propose the use of LLM for disfluency rephrasing. However, this method typically involve the fine-tuning of LLM, incurring significant costs in terms of resources and efforts. Nevertheless, the direct use of LLM in practical applications is challenging due to high memory usage and computational intensity. This has inspired our investigation into the generation of disfluent sentences without the need for fine-tuning LLM. Moreover, there are still challenges associated with using LLMs to generate disfluent sentences: First, how to design effective prompts to instruct LLM in generating disfluent sentences? Secondly, the generated disfluent sentences may exhibit poor quality, deviating from real-world scenarios or being overly simplistic.

To solve these problems, in this work, we introduce the use of LLM's powerful generation capabilities and natural language understanding to generate disfluent sentences as augmentation data for training a small disfluency detection model. This approach distills the knowledge from LLM into small detector models, resulting in efficient and lightweight models. Unlike previous efforts in disfluency generation, our approach eliminates the necessity for fine-tuning LLM. This makes it cost-effective and highly efficient, leading to excellent results even with a small amount of generated data.

To summarize, our contributions as follows: 
\begin{itemize}
\item We propose a framework that addresses data sparsity issues by generating disfluent data using LLM as augmentation data. This approach allows us to distill knowledge into a small model, achieving a lightweight and efficient disfluency detection model.
\item We design specific prompts instead of fine-tuning LLM to generate annotated disfluent sentences. Compared to other methods, our generation approach is cost-effective, and the generated disfluent sentences are more realistic and diverse.
\item To achieve better data augmentation effects, we introduce a data filtering step and conduct comprehensive experiments using the filtered data to evaluate the effectiveness of our approach.
\end{itemize}

Notably, our method achieves state-of-the-art results using a limited amount of disfluent sentences, which ensures greater efficiency and cost-effectiveness.

\section{Related Work}
\subsection{Disfluency Detection}
The landscape of disfluency detection models falls into five main categories: sequence tagging, noisy channel, translation-based, parsing-based, and data augmentation methods. Sequence tagging treats disfluency detection as a sequence labeling task, with Lin et al. \cite{lin20c_interspeech} employing an attention-based structure in the multi-task learning framework, and Zayats et al. \cite{zayats2019giving} utilizing text-based distributional prediction of acoustic cues for vector z-score features. Rocholl et al. \cite{rocholl2021disfluency} proposed on-device lightweight BERT architectures. Lou et al. \cite{jamshid-lou-etal-2018-disfluency} introduced a new auto-correlational kernel, and Wang et al. \cite{wang2016neural} employed RNN. The noisy channel model, introduced by Lou et al. \cite{jamshid-lou-johnson-2017-disfluency}, treats disfluencies as errors during message encoding and decoding, then utilizes a long short-term memory neural network language model to rescore candidate disfluency analyses. Translation-based approaches formulate disfluency detection as encoder-decoder systems. Dong et al. \cite{dong2019adapting} adapted the neural machine translation model, and Wang et al. \cite{wang-etal-2018-semi} extended the traditional encoder-decoder model by leveraging two independent encoders. In the parsing-based approach, as explored by Sen et al. \cite{sen-groves-2021-semantic}, semantic parsing is employed to identify disfluency. Tran et al. \cite{tran-etal-2018-parsing} utilized acoustic-prosodic features extracted from the text. To address data scarcity and reduce reliance on gold-standard annotated data, researchers have proposed data augmentation techniques. Rocholl et al. \cite{rocholl2021disfluency} emploted Wikipedia+Books for model pretraining.
\begin{figure}[t]
\centering 
\includegraphics[scale=0.3]{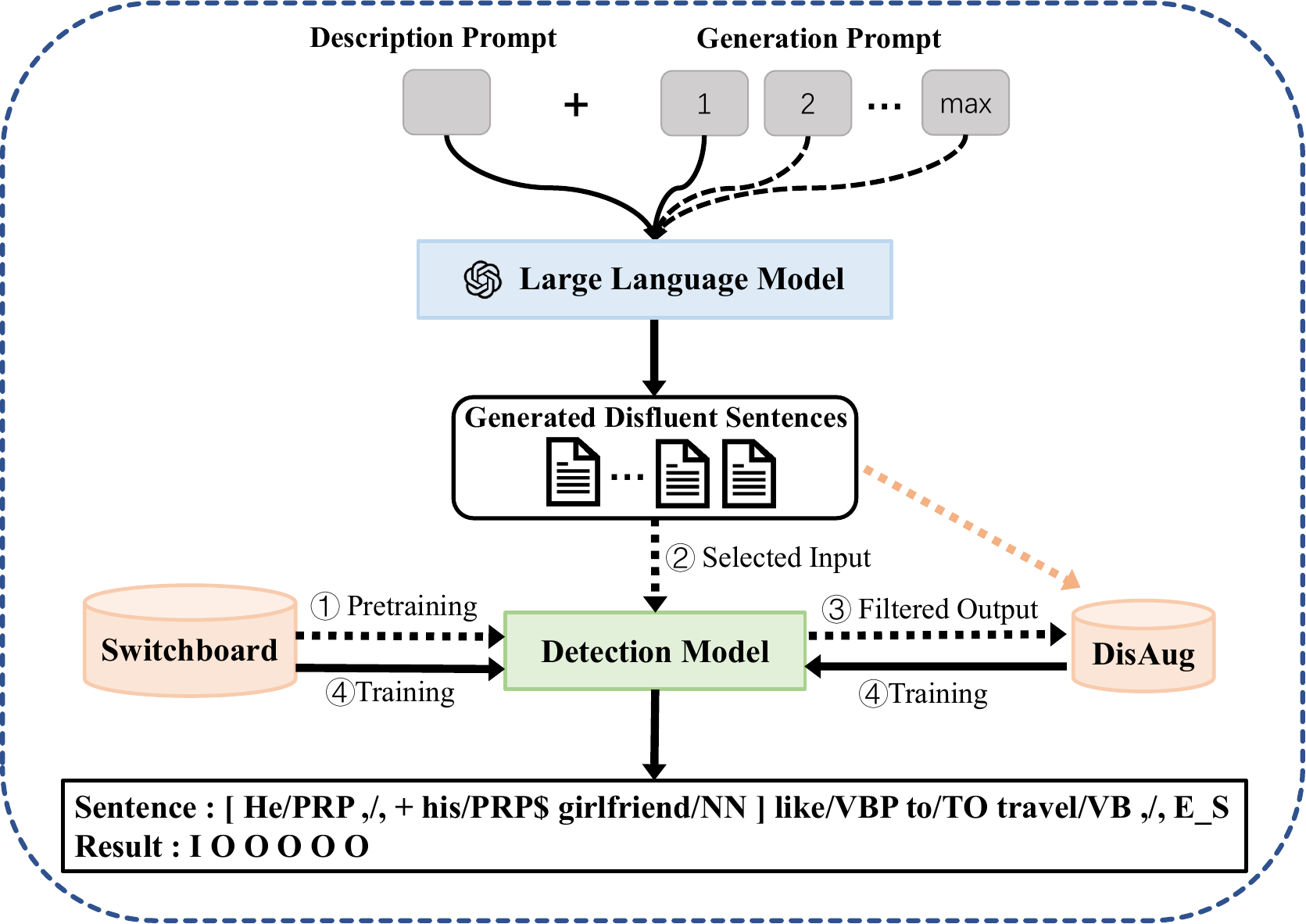}
\caption{An Overview of Our Approach: For disfluency generation, we begin by providing a description prompt to the LLM and guide its generation of disfluent sentences through a series of generation prompts. Subsequently, we use the trainable detection model to filter out high-quality sentences after pretraining on Switchboard, namely DisAug. For disfluency detection, DisAug is incorporated with Switchboard as training data to train the original detection model.}
\label{overall}
\end{figure}
\subsection{Disfluency Generation}
Passali et al. \cite{passali-etal-2022-lard} used heuristic rules to generate disfluenct sentences. Gupta et al. \cite{gupta2021disfl} built DISFL-QA, a benchmark dataset of contextual disfluency in QA through manual annotation. Wang et al. \cite{wang2020multi} and Wang et al. \cite{wang-etal-2020-combining} generated disfluenct sentences by randomly repeatting, injecting and deleting n-grams from unlabeled news data as additional training data. However, their approaches are admittedly generating less natural disfluent sentences than with a neural model. Yang et al. \cite{yang-etal-2020-planning} introduced the PGmodel, a state-of-the-art data augmentation model, to generate disfluency, in which a Planner determines the appropriate placement of disfluent segments, and a Generator creates the corresponding disfluent segments. Marie et al. \cite{marie2023disfluency} proposed a disfluency paraphraser, utilizing LLM to rephrase fluent sentences into disfluent ones. While their approaches can generate natural disfluent sentences, but it requires training an additional generation model, leading to higher costs for fine-tuning LLM.

\section{Methodology}
In this section, we first define the problem of disfluency detection and introduce the detection model. Next, we describe the prompt designs guiding the Language Model (LLM) to generate disfluent sentences as augmented data, as well as the data filters used to select high-quality generated sentences. The overall framework is illustrated in Fig. \ref{overall}.

\subsection{Task Definition}
We consider the disfluency detection task as a sequence labeling task. The model takes as input a dataset containing $N$ sentences $\left\{S_1, S_2, ..., S_N\right\}$, where the $i$-th sentence is represented as $s_i = \left\{w_1,w_2,...,w_T\right\}$, with $T$ being the number of words in that sentence. Our objective is to output a label sequence for each sentence, denoted as $Y_i = \left\{y_1,y_2,...,y_T \right\}$, where $I (O)$ indicates whether the word is inside (or outside) the reparandum region.
\subsection{Disfluency Generation}
\subsubsection{Data Generation}
The LLM demonstrates a profound comprehension of context and semantics, highlighting its remarkable ability to generate coherent and fluent sentences. However, when it comes to generating sentences with disfluencies, an area where LLM faces challenges, it cannot be directly instructed as in the case of generating fluent sentences tasks. Our goal is to enable the LLM to generate disfluent sentences using a simple and efficient approach. Therefore, we propose a Description-then-Generation disfluency generation method. First, we employ a data description prompt for the LLM. This prompt conveys data descriptions as instructions, enabling the LLM to understand the intended data for generation, along with its concepts and features. Next, we input generation prompts into the LLM. This further enhances the LLM's understanding of the desired generated data, facilitating the creation of diverse disfluent sentences. Given the LLM's extensive contextual understanding, we set the maximum number of generation round as \textit{max round}, requiring the reintroduction of the Description Prompt after every \textit{max round} rounds of generation. 
\begin{textbox}{Description Prompt.}
The following is the description of disfluenct sentence:
1. Labeling Format\\
Filled pause: \{F ...\}; Explicit editing term: \{E ...\}; Discourse marker: \{D ...\}; Coordinating conjunction: \{C ...\}; Aside: \{A ...\}.

2. Part-of-speech Tags Format\\
POS tagging assigns grammatical labels to words in a sentence, such as NN for nouns, VB for verbs, JJ for adjectives, RB for adverbs, PRP for pronouns, IN for prepositions, CC for conjunctions, DT for determiners, UH for interjections, and CD for numbers.

3. Disfluency Types \\
Repetition: certain words or phrases are repeated; Substitution: replacement of the preceding word or phrase; Deletion: removal of specific words; Chaining: continuity where something from each part carries over into the next; Nesting: a restart within a restart.

4. These disfluency types can combine to form more complex sentences. The `[ ]' denotes the disfluent part, and the `+' symbol indicates the end of reparandum.
\end{textbox}
\begin{textbox}{Generation Prompt.}
These example sentences contain part-of-speech tags and disfluency annotation. Please generate more sentences which are similar with example sentences. The generated sentences also have part-of-speech tags and disfluency annotation. Just give me answers, no more tips words and don't copy the example sentences.
Note:  generated sentences need to belong various disfluency types such as restart, delete, substitution.  Two or Three Example sentences: (Providing examples of disfluent sentence)
\end{textbox}

\renewcommand{\algorithmicrequire}{\textbf{Input:}}  
\renewcommand{\algorithmicensure}{\textbf{Output:}} 
\begin{algorithm}[h]
\small
  \caption{Disfluent Sentence Generation Algorithm} 
  \label{algo}
  \begin{algorithmic}[1]
    \Require
      Prompt 1: Description prompt;
      Prompt 2: Generation prompt;
    \Ensure
       Generated disfluent sentences $S_{dis}$;
       \State set $i=0$, Maximum number of generation round $max\_round$ ;
		
        \State Input Prompt 1 to the LLM;

        \While{ $i \leq max\_round$}
        
        \State Input Prompt 2 to the LLM;
        \State Get generated sentences-$>S_{dis}$
        \State $i++$;
		
	\EndWhile
    \State Return $S_{dis}$
  \end{algorithmic}
\end{algorithm}
\vspace{-1em}
It is worth noting that the example sentences in the generation prompt not only contain disfluency annotations but also include part-of-speech tags. This inclusion enables the LLM to better discern the various types of disfluent words and disfluent grammatical structures. We can specify the types of disfluencies present in the example sentences to control the disfluency types generated in the resulting sentences. Besides, the inclusion of example sentences contributes to making the sentences generated by the LLM more human-like.
\subsubsection{Uncertainty-aware Data Filtering}
Considering that LLM tends to bias towards generating fluent sentences, leading to a decline in the quality of disfluent sentences, and aiming to achieve higher performance gains with minimal augmented data, inspired by graph collaborative filtering\cite{guo2023manipulating}, we introduce an uncertainty-aware data filtering. This process identifies high-confidence disfluent sentences and excludes those potentially detrimental to model performance.

The uncertainty measure is computed by averaging the predicted probabilities of words labeled as disfluent. The higher the predicted probability for disfluent labels, the lower the uncertainty of the disfluent sentence. Therefore, a value closer to 1 indicates higher sentence quality. The computation formula is presented in Eq. \ref{uncertainty}, where $p_j$ represents the predicted probability of the j-th word being disfluent, and $y_j$ is the label associated with the j-th word, indicating whether it is disfluent (labeled as `I'). We set the hyperparameter $\lambda$ as the minimum threshold. If the computed average predicted probability falls below this predefined threshold, it indicates that the sentence is considered to have high uncertainty in its disfluent portions. Therefore, such sentences will be filtered out or excluded from further consideration.
\begin{equation}
\label{uncertainty}
    \frac{1}{K}\sum_{j=1}^{K} p_j,\forall p_j\vDash y_j \in \{I\}
\end{equation}
\subsubsection{Model Agnostic Disfluency Detection}

There are many models available for sequence labeling tasks of disfluency detection. We aim to use data augmentation to enable simple models to perform better. Therefore, we directly chose trainable language models as detectors for pretraining and fine-tuning. Specifically, we obtain the hidden state $h_i$ for each word $w_i$, and then calculate the label probabilities $p_i$ for each word, as shown in Eq. \ref{bert}.
\begin{equation}
    \begin{split}
    H={h_1,h_2,...,h_T}&={\rm Detector}({w_1,w_2,...,w_T})
    \\
    p_i&={\rm softmax}(Wh_i+b)
    \label{bert}
    \end{split}
\end{equation}

The objective of this model is to minimize the cross-entropy loss:
\begin{equation}
    L = \mathbb{E}_{s,l}\sum_{i=1}^{T}{\rm CrossEntropy}(l_i,p_i)    
\end{equation}
where $l_i$ and $p_t$ represent the true label predicted probability distribution of the $i$-th word respectively. $\mathbb{E}_{s,l}$ represents the expected value of all possible labeling sequences.
\begin{table}[t]
  \centering 
  \caption{Generated Disfluent Sentences. The data annotation includes part-of-speech tagging and disfluency labeling. The underlined part represents the disfluent portion in the original sentence, and the boldface represents the disfluent portion in the generated sentence.}
    \resizebox{\linewidth}{!}{\begin{tabular}{ll}
    \toprule  
    
    \textbf{Original Sentence} 
    & 
    \textbf{Generated Disfluent Sentence} \\ 
    
    \midrule
    \multicolumn{2}{l}{\textbf{Heuristic Method}}\\
    \midrule
    
    \makecell[l]{Sometimes/RB ,/, \underline{\{F uh/UH ,/, \}} it/PRP 's/BES\\a/DT bit/NN of/IN \underline{{[} a/DT ,/, + a/DT {]}} problem/NN ,/,\\\underline{\{D you/PRP know/VBP ,/, \}} because/IN I/PRP guess/VBP\\I/PRP do/VBP n't/RB really/RB manage/VB my/PRP\$\\money/NN the/DT way/NN I/PRP should/MD ./. E\_S }
    &
    \makecell[l]{At/IN times/NNS ,/, \underline{\{F uh/UH ,/, \}} \textbf{{[} it/PRP ,/, +}\\\textbf{it/PRP {]}} can/MD be/VB \textbf{{[} a/DT little/JJ ,/, + a/DT}\\\textbf{little/JJ {]}}problematic/JJ ,/, \underline{\{D you/PRP know/VBP ,/, \}}\\because/IN I/PRP suppose/VBP I/PRP do/VBP n't/RB\\ quite/RB manage/VB my/PRP\$ money/NN as/IN\\effectively/RBas/IN I/PRP should/MD ./.  E\_S} 
    \\
    \midrule
\makecell[l]{\underline{\{C and/CC \}} they/PRP are/VBP beginning/VBG\\to/TO be/VB a/DT budget/NN problem/NN but/CC ,/,\\\underline{\{F uh/UH ,/, \}} have/VBP not/RB been/VBN really/RB\\\underline{{[} up/IN until/IN this/DT ,/, + up/IN to/IN this/DT {]}}\\point/NN ./. E\_S}
    &
    \makecell[l]{\underline{\{C and/CC \}} they/PRP are/VBP becoming/VBG a/DT\\budget/NN issue/NN but/CC ,/, \underline{\{F uh/UH ,/, \}} have/VBP\\n't/RB really/RB \textbf{{[} caught/VBN up/IN until/IN this/DT}\\\textbf{,/, + caught/VBN up/IN to/IN this/DT {]}} point/NN ./. E\_S}
    \\
    \midrule
    \makecell[l]{Now/UH I/PRP know/VBP my/PRP\$ boss/NN \\has/VBZ bought/VBN the/DT software/NN ,/, \\\underline{\{F um/UH ,/, \} {[} that/WDT he/PRP can/MD ,/, }\\\underline{+ that/WDT  his/PRP\$ {]}} checkbook/NN is/VBZ\\\underline{{[} on/IN ,/, + on/IN {]} + on/IN {]}} a/DT disk/NN ,/, E\_S}
    &
    \makecell[l]{Now/UH \textbf{{[} I/PRP realize/VBP ,/, + \{F um/UH ,/, \} }\\\textbf{I/PRP understand/VBP {]} }that/DT my/PRP\$ boss/NN\\has/VBZ procured/VBN the/DT software/NN ,/, \\\textbf{{[} indicating/VBG that/WDT he/PRP can/MD ,/, +}\\\textbf{allowing/VBG him/PRP\$ to/TO {]}}store/VB his/PRP\$\\checkbook/NN details/NNS on/IN a/DT disk/NN ,/, E\_S}
    \\
    \bottomrule
\end{tabular}}
  \label{generated-sentence}%
\end{table}

\begin{figure}[tbp]

\centering 

\includegraphics[scale=0.28]{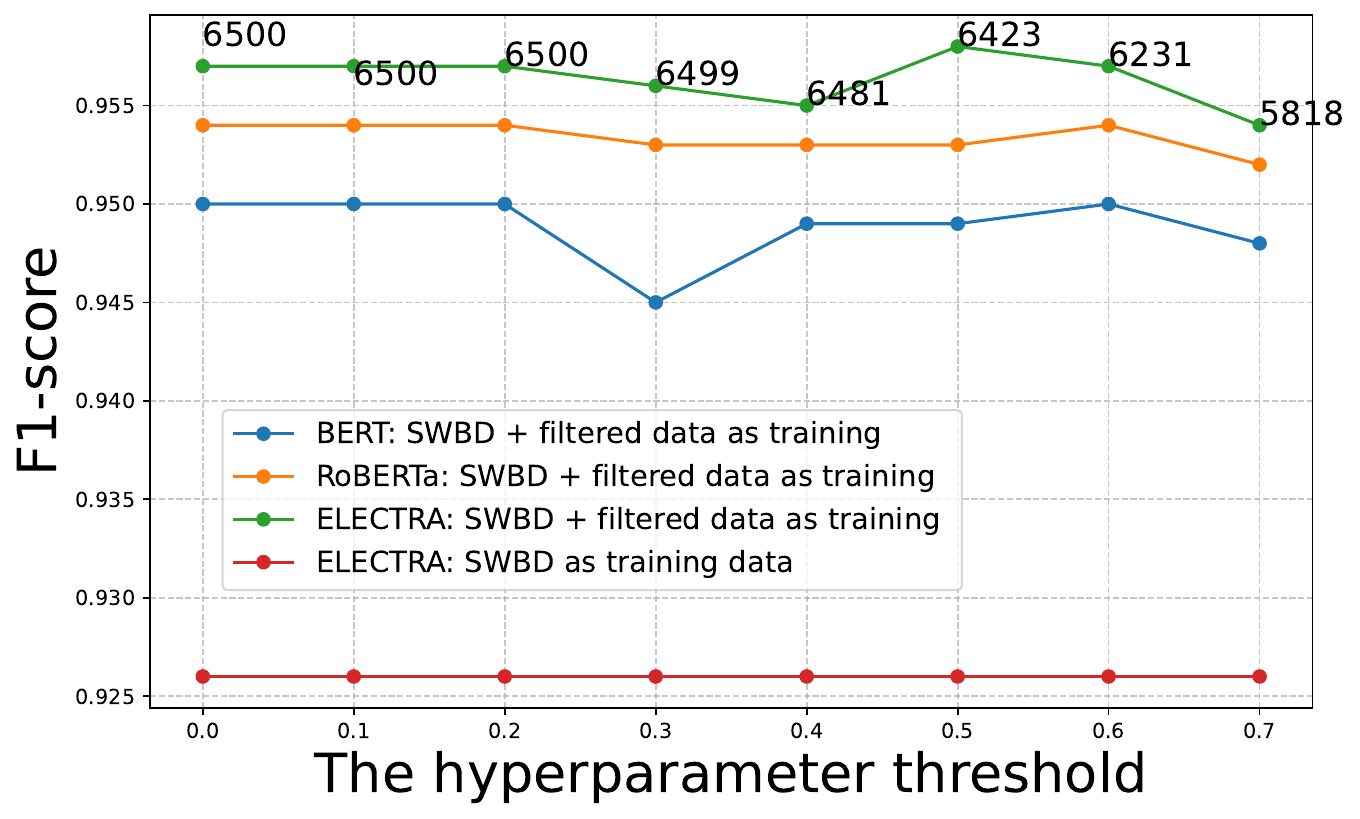}

\caption{F1-scorce under different confidence thresholds. The values on the line represent the quantity of generated sentences filtered at that threshold.}
\label{lamuda}
\end{figure}
\section{Experiments}
\subsection{Experiment Setup}
\textbf{Dataset.}\quad We choose Switchboard (SWBD) \cite{godfrey1992switchboard}, the standard and largest disfluency detection benchmark dataset for evaluating the proposed method. Following the experiment setup outlined in Charniak and Johnson\cite{charniak-johnson-2001-edit}, we partitioned the Switchboard corpus into training, development, and test sets. The training set comprises 173k sentences, the development set includes 10k sentences, and the test set encompasses 7.9k sentences.

 \textbf{Baselines.}\quad Transformers \cite{wang2020multi} and Transformers \& BERT \cite{wang2020multi} employed heuristic methods to generate disfluent data for pretraining, while BERT-CRF \cite{lee2020auxiliary} and ELECTRA-CRF \cite{lee2020auxiliary} adopted a multitask approach. BERT-GCN \cite{ghosh2022span} leveraged structural information for span-level detection, which is currently the SOTA method for the disfluency detection task. And we compare the PGmodel \cite{yang-etal-2020-planning} (i.e., BERT-GRI3M model and BERT-GRI20M model) and other models that performed disfluency detection using different data augmentation techniques. Marie et al. \cite{marie2023disfluency} fine-tuned T5 to rewrite fluent sentences as disfluent ones, generating data for training a disfluency classifier. As the generated data has not been disclosed, we employe the disfluency classifier as the detector for experimental comparisons. In addition, we employ three disfluency detectors, namely bert-base-uncased \cite{DBLP:journals/corr/abs-1810-04805}, roberta-base\cite{DBLP:journals/corr/abs-1907-11692}, and google/electra-base-discriminator \cite{clark2020electra}, using their fine-tuning outcomes on SWBD as baselines for comparison. In order to comprehensively examine the effect of the data enhancement we generated, we also use DISFL-QA \cite{gupta2021disfl} as the augmented data for experimental comparison.

\begin{table}[t]
\tiny
    \centering    
    \caption{Percentage of generated disfluency sentences.}
    \resizebox{6cm}{!}{
    \begin{tabular}{lccc}
    \toprule
       \textbf{Generation Model}& \textbf{Repetition} & \textbf{Substitution} & \textbf{Deletion} \\
       \midrule
       PG\cite{yang-etal-2020-planning} & 85.68\%&13.08\% &1.24\%
       \\
       PG-NC-AD-ID\cite{yang-etal-2020-planning} &19.80\% &25.27\% &54.93\%
       \\
       DisAug &34.57\% &44.10\% &21.32\%
       \\
       \bottomrule
    \end{tabular}}
    
    \label{percentage}
\end{table}

\begin{table*}[ht]
\tiny
  \centering 
\caption{Results of disfluency detection. The \dag mark denotes that the results are significant with the significance level $p < 0.05$. F1-score is the major metric.}  
\resizebox{\linewidth}{!}{
\begin{tabular}{c|c|ccc}
\hline
\multicolumn{5}{c}{\multirow{2}{*}{\textbf{Baselines}}}\\
\multicolumn{5}{c}{}\\
\hline
\textbf{Model} & \textbf{Setting} & \textbf{Precision} & \textbf{Recall} & \textbf{F1-score} \\ \hline
Transformers\cite{wang2020multi}         & multitask \& 3M pretraining & 0.934 & 0.873 & 0.902 \\
Transformers \& BERT\cite{wang2020multi} & multitask \& 3M pretraining & -     & -     & 0.914 
\\
BERT-CRF\cite{lee2020auxiliary} & multitask& 0.946& 0.912&0.929 \\
ELECTRA-CRF\cite{lee2020auxiliary}& multitask& 0.948&0.916&0.931 \\
BERT-GCN\cite{ghosh2022span}& Span Classification  & \underline{0.952} & \underline{0.932}& \underline{0.942}\\
\hline
BERT-GRI3M(PGmodel)\cite{yang-etal-2020-planning}  & 3M pretraining              & 0.951 & 0.894 & 0.922 \\
BERT-GRI20M(PGmodel)\cite{yang-etal-2020-planning} & 20M pretraining             & 0.945 & 0.902 &0.923 \\

\hline

Disfluency Classifier& \multirow{4}{*}{SWBD as training data (173K)}& 0.944 & 0.842 & 0.890 \\
BERT& &0.949 & 0.867 & 0.906 \\
RoBERTa& & 0.946&0.883 &0.913 \\
ELECTRA& &0.939 &0.914 &0.926 \\
 
\hline
ELECTRA&SWBD + DISFL-QA (7.1K) as training data& 0.940&0.906&0.923 \\
\hline
\multicolumn{5}{c}{\multirow{2}{*}{\textbf{Ours}}}\\
\multicolumn{5}{c}{}\\
\hline
\textbf{Model} & \textbf{Setting} & \textbf{Precision} & \textbf{Recall} & \textbf{F1-score} \\ \hline
\multirow{4}{*}{Disfluency Classifier}  
                 &        DisAug + FluData as training data (8K)       
      &0.944&0.880&0.742
 \\ 

 &  DisAug pretraining            &0.934       &0.856       &0.893       
 \\
    &   SWBD + Original Data as training data            
    &\textbf{0.956}&0.925&0.940
    \\
        
 &SWBD + DisAug (6.4K) as training data&0.943 & \textbf{0.944} &\textbf{0.944}\\
\hline
\multirow{4}{*}{BERT}        
                 &        DisAug + FluData as training data (8K)       
      &0.903&0.920&0.912
 \\ 

 &  DisAug pretraining            &0.954       &0.895       &0.923
 \\
    &   SWBD + Original Data as training data            
    &0.954&\textbf{0.947}&\textbf{0.950}
    \\
        
 &SWBD + DisAug (6.4K) as training data&0.956 & 0.942 &0.949\\
\hline
\multirow{4}{*}{RoBERTa} 
& DisAug + FluData as training data (8K)
  &0.947&0.882&0.914
       \\
& DisAug pretraining
& 0.949 &0.894  &0.920 
       \\
       & SWBD + Original Data as training data
       &\textbf{0.957}&0.949&0.953
       \\
       &SWBD + DisAug (6.4K) as training data &0.947 &\textbf{0.961} &\textbf{0.954}\\
       \hline
\multirow{4}{*}{ELECTRA} 
&DisAug + FluData as training data (8K) 
 &0.918&0.945&0.931\\

&DisAug pretraining  
&0.947&0.927&0.937\\
 & SWBD + Original Data as training data
 &0.955&0.959&0.957\\
&SWBD + DisAug (6.4K) as training data &\textbf{0.955} &\textbf{0.960} &\textbf{0.958}\dag\\

\hline
\end{tabular}}

\label{table4}
\end{table*}

\textbf{Implementation Details.}
We use OpenAI’s text-davinci-003 language model as the disfluency generation model. For disfluency detection, we utilize the Adam optimizer with a learning rate of 1e-5 and batch size 32 during the pretraining and fine-tuning stages. And, we uniformly employe the ELECTRA model fine-tuned on SWBD for data filtering, with the confidence threshold hyperparameter $\lambda$ set to 0.5.
\subsection{Experimental Results}

\subsubsection{Disfluency Generation Results}
We generated 6.5K original disfluent sentences, and then pretrained the ELECTRA model using SWBD as a data filter, resulting in 6423 selected sentences, which we named `DisAug'. Table \ref{generated-sentence} illustrates the comparison between DisAug and sentences generated by other methods. DisAug serves as our augmented data for disfluency detection. In Table \ref{percentage}, we showcase the percentage distribution of three types of disfluencies observed in the generated dataset: repetition, substitution, and deletion. PG \cite{yang-etal-2020-planning} and PG-NC-AD-ID \cite{yang-etal-2020-planning} refer to two different generation methods employed in the PG model. It's worth noting that many sentences in DisAug exhibit intricate patterns. In comparison to disfluent sentences generated through heuristic-based and PG model methods, sentences in DisAug often manifest more than one type of disfluency and tend to feature complex patterns of substitution.

Figure \ref{lamuda} shows the number of disfluent sentences filtered by different label confidence thresholds and the results as augmented data to enhance disfluency detection. We can observe that the enhancement effect on disfluency detection shows little fluctuation between different filtered generated sentences. This suggests that the sentences generated using our method exhibit stability and efficiency, thereby improving the model's performance in disfluency detection while reducing the workload for data generation.

\begin{figure}[tbp]

\centering 
\includegraphics[scale=0.25]{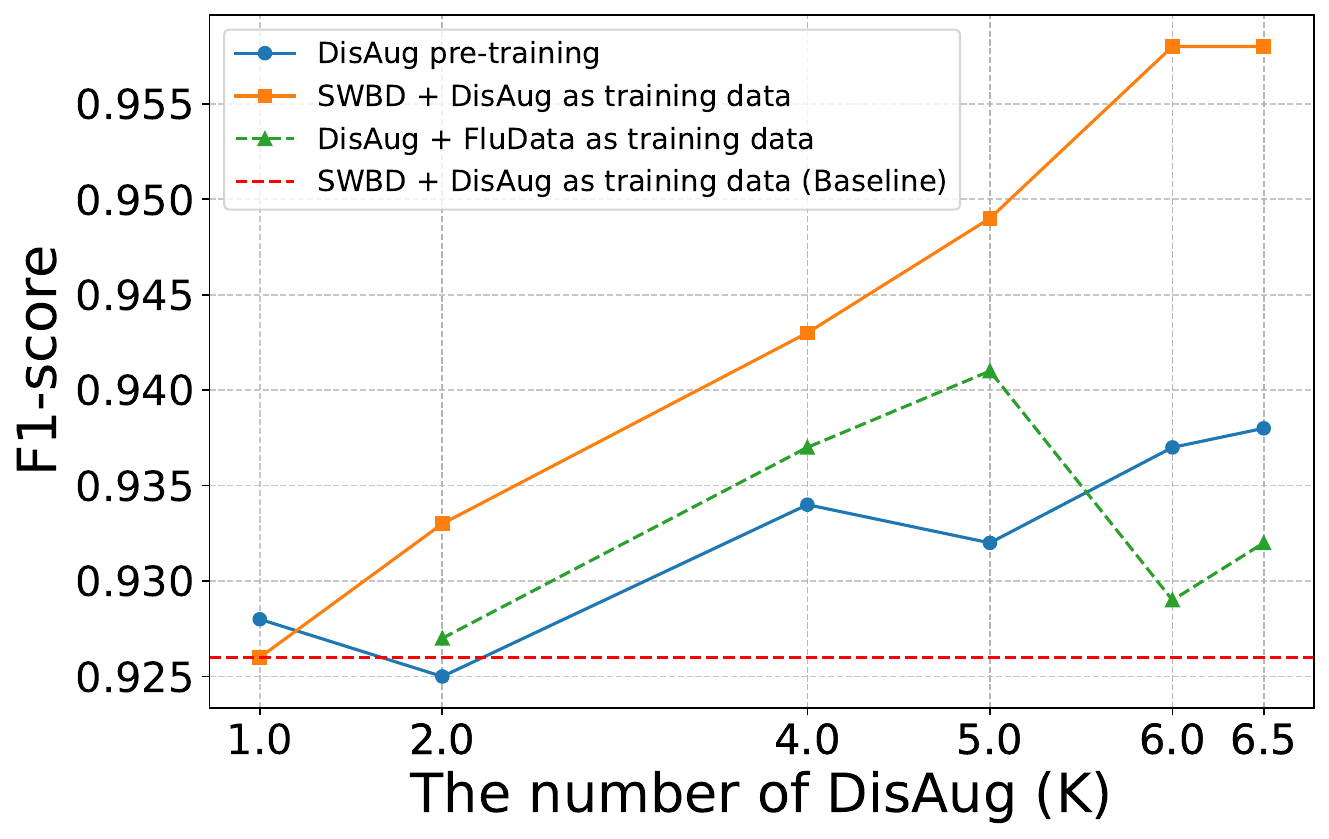}
\caption{Results of Disfluency Detection with ELECTRA as a Detector under Different Experimental Setups.}
\label{photo}
\end{figure}

\subsubsection{Disfluency Detection Results}
To further validate the effectiveness of the generated disfluent data and improve the performance of disfluency detection, we designed four setups of experiments: (1) Combing DisAug with FluData (fluent sentences generated by LLM) as the training set for fine-tuning the model; (2) Pretraining the language model using DisAug and then fine-tuning it on the SWBD training set; (3) Adding original generated data to the SWBD training set as training data for fine-tuning the model; (4) Adding DisAug to the SWBD training set as training data for fine-tuning the model. In these four experimental setups, the dev set and test set remained consistent and unchanged. The main experimental results are presented in Table \ref{table4}. Figure \ref{photo} illustrates the variation in the F1-score of the ELECTRA detector for disfluency detection when different quantities of generated data are employed across various experiments. 

From the experimental results, we observed a significant improvement in the model's performance with the generated augmented data. Incorporating DisAug into SWBD for training the ELECTRA model achieved SOTA results. Compared to the manually annotated DISFL-QA dataset, our approach has achieved superior results, demonstrating the efficiency and cost-effectiveness of our method. We exceeded the majority of baselines by using 8K generated sentences as training data, effectively reducing training costs. Additionally, by adjusting the proportion of generated disfluent and fluent sentences, we can further enhance the disfluency detection performance. The original data and DisAug yielded competitive results, which indirectly verified the effectiveness of our data generation method.
\section{CONCLUSION}
In this study, we propose a method that guides LLM with specific prompts to generate more diverse and natural disfluent sentences, thereby enhancing disfluency detection. We also implement a data filtering process to improve data quality. Experimental results consistently affirm the efficiency and cost-effectiveness of our approach, demonstrating its effectiveness with limited data resources. In the future, we plan to explore the integration of speaker characteristics to deploy our method in online environments. We will consider extending its application to other tasks, such as disfluent paraphrasing.

\bibliographystyle{IEEEbib}
\bibliography{ref}

\begin{thebibliography}{10}

\bibitem{futami2023streaming}
Hayato Futami, Emiru Tsunoo, Kentaro Shibata, Yosuke Kashiwagi, Takao Okuda, Siddhant Arora, and Shinji Watanabe,
\newblock ``Streaming joint speech recognition and disfluency detection,''
\newblock in {\em ICASSP 2023-2023}. IEEE, 2023, pp. 1--5.

\bibitem{marie2023disfluency}
Benjamin Marie,
\newblock ``Disfluency generation for more robust dialogue systems,''
\newblock in {\em Findings of the Association for Computational Linguistics: ACL 2023}, 2023, pp. 11479--11488.

\bibitem{dinkar2023fillers}
Tanvi Dinkar, Chlo{\'e} Clavel, and Ioana Vasilescu,
\newblock ``Fillers in spoken language understanding: Computational and psycholinguistic perspectives,''
\newblock {\em arXiv preprint arXiv:2301.10761}, 2023.

\bibitem{wang-etal-2020-combining}
Shaolei Wang, Zhongyuan Wang, Wanxiang Che, and Ting Liu,
\newblock ``Combining self-training and self-supervised learning for unsupervised disfluency detection,''
\newblock in {\em EMNLP}, Bonnie Webber, Trevor Cohn, Yulan He, and Yang Liu, Eds., Online, Nov. 2020, pp. 1813--1822, Association for Computational Linguistics.

\bibitem{shriberg1997prosody}
Elizabeth Shriberg, Rebecca~A. Bates, and Andreas Stolcke,
\newblock ``A prosody only decision-tree model for disfluency detection,''
\newblock in {\em EUROSPEECH}, 1997.

\bibitem{jamshid-lou-johnson-2020-improving}
Paria Jamshid~Lou and Mark Johnson,
\newblock ``Improving disfluency detection by self-training a self-attentive model,''
\newblock in {\em ACL}, Dan Jurafsky, Joyce Chai, Natalie Schluter, and Joel Tetreault, Eds., Online, July 2020, pp. 3754--3763, Association for Computational Linguistics.

\bibitem{wang2020multi}
Shaolei Wang, Wangxiang Che, Qi~Liu, Pengda Qin, Ting Liu, and William~Yang Wang,
\newblock ``Multi-task self-supervised learning for disfluency detection,''
\newblock in {\em Proceedings of the AAAI Conference on Artificial Intelligence}, 2020, vol.~34, pp. 9193--9200.

\bibitem{rocholl2021disfluency}
Johann~C Rocholl, Vicky Zayats, Daniel~D Walker, Noah~B Murad, Aaron Schneider, and Daniel~J Liebling,
\newblock ``Disfluency detection with unlabeled data and small bert models,''
\newblock {\em arXiv preprint arXiv:2104.10769}, 2021.

\bibitem{passali-etal-2022-lard}
Tatiana Passali, Thanassis Mavropoulos, Grigorios Tsoumakas, Georgios Meditskos, and Stefanos Vrochidis,
\newblock ``{LARD}: Large-scale artificial disfluency generation,''
\newblock in {\em LREC}, Nicoletta Calzolari, Fr{\'e}d{\'e}ric B{\'e}chet, Philippe Blache, Khalid Choukri, Christopher Cieri, Thierry Declerck, Sara Goggi, Hitoshi Isahara, Bente Maegaard, Joseph Mariani, H{\'e}l{\`e}ne Mazo, Jan Odijk, and Stelios Piperidis, Eds., Marseille, France, June 2022, pp. 2327--2336, European Language Resources Association.

\bibitem{yang-etal-2020-planning}
Jingfeng Yang, Diyi Yang, and Zhaoran Ma,
\newblock ``Planning and generating natural and diverse disfluent texts as augmentation for disfluency detection,''
\newblock in {\em EMNLP}, 2020, pp. 1450--1460.

\bibitem{zhao2023survey}
Wayne~Xin Zhao, Kun Zhou, Junyi Li, Tianyi Tang, Xiaolei Wang, Yupeng Hou, Yingqian Min, Beichen Zhang, Junjie Zhang, Zican Dong, et~al.,
\newblock ``A survey of large language models,''
\newblock {\em arXiv preprint arXiv:2303.18223}, 2023.

\bibitem{sun2023towards}
Hao Sun, Hengyi Cai, Bo~Wang, Yingyan Hou, Xiaochi Wei, Shuaiqiang Wang, Yan Zhang, and Dawei Yin,
\newblock ``Towards verifiable text generation with evolving memory and self-reflection,''
\newblock {\em arXiv preprint arXiv:2312.09075}, 2023.

\bibitem{sun2023allies}
Hao Sun, Xiao Liu, Yeyun Gong, Yan Zhang, and Nan Duan,
\newblock ``Allies: Prompting large language model with beam search,''
\newblock in {\em Findings of EMNLP 2023}, 2023.

\bibitem{lin20c_interspeech}
Binghuai Lin and Liyuan Wang,
\newblock ``Joint prediction of punctuation and disfluency in speech transcripts.,''
\newblock in {\em Interspeech}, 2020, pp. 716--720.

\bibitem{zayats2019giving}
Vicky Zayats and Mari Ostendorf,
\newblock ``Giving attention to the unexpected: Using prosody innovations in disfluency detection,''
\newblock {\em arXiv preprint arXiv:1904.04388}, 2019.

\bibitem{jamshid-lou-etal-2018-disfluency}
Paria Jamshid~Lou, Peter Anderson, and Mark Johnson,
\newblock ``Disfluency detection using auto-correlational neural networks,''
\newblock in {\em EMNLP}. 2018, pp. 4610--4619, Association for Computational Linguistics.

\bibitem{wang2016neural}
Shaolei Wang, Wanxiang Che, and Ting Liu,
\newblock ``A neural attention model for disfluency detection,''
\newblock in {\em COLING}, 2016, pp. 278--287.

\bibitem{jamshid-lou-johnson-2017-disfluency}
Paria Jamshid~Lou and Mark Johnson,
\newblock ``Disfluency detection using a noisy channel model and a deep neural language model,''
\newblock in {\em Proceedings of the 55th Annual Meeting of the Association for Computational Linguistics (Volume 2: Short Papers)}, Vancouver, Canada, July 2017, pp. 547--553, Association for Computational Linguistics.

\bibitem{dong2019adapting}
Qianqian Dong, Feng Wang, Zhen Yang, Wei Chen, Shuang Xu, and Bo~Xu,
\newblock ``Adapting translation models for transcript disfluency detection,''
\newblock in {\em AAAI}, 2019, vol.~33, pp. 6351--6358.

\bibitem{wang-etal-2018-semi}
Feng Wang, Wei Chen, Zhen Yang, Qianqian Dong, Shuang Xu, and Bo~Xu,
\newblock ``Semi-supervised disfluency detection,''
\newblock in {\em COLING}, 2018, pp. 3529--3538.

\bibitem{sen-groves-2021-semantic}
Priyanka Sen and Isabel Groves,
\newblock ``Semantic parsing of disfluent speech,''
\newblock in {\em Proceedings of the 16th Conference of the European Chapter of the Association for Computational Linguistics: Main Volume}, Paola Merlo, Jorg Tiedemann, and Reut Tsarfaty, Eds., Online, Apr. 2021, pp. 1748--1753, Association for Computational Linguistics.

\bibitem{tran-etal-2018-parsing}
Trang Tran, Shubham Toshniwal, Mohit Bansal, Kevin Gimpel, Karen Livescu, and Mari Ostendorf,
\newblock ``Parsing speech: a neural approach to integrating lexical and acoustic-prosodic information,''
\newblock in {\em NAACL-HLT 2018}, pp. 69--81.

\bibitem{gupta2021disfl}
Aditya Gupta, Jiacheng Xu, Shyam Upadhyay, Diyi Yang, and Manaal Faruqui,
\newblock ``Disfl-qa: A benchmark dataset for understanding disfluencies in question answering,''
\newblock {\em arXiv preprint arXiv:2106.04016}, 2021.

\bibitem{guo2023manipulating}
Jiayan Guo, Lun Du, Xu~Chen, Xiaojun Ma, Qiang Fu, Shi Han, Dongmei Zhang, and Yan Zhang,
\newblock ``On manipulating signals of user-item graph: A jacobi polynomial-based graph collaborative filtering,''
\newblock in {\em Proceedings of the 29th ACM SIGKDD Conference on Knowledge Discovery and Data Mining}, 2023, pp. 602--613.

\bibitem{godfrey1992switchboard}
John~J Godfrey, Edward~C Holliman, and Jane McDaniel,
\newblock ``Switchboard: Telephone speech corpus for research and development,''
\newblock in {\em Acoustics, speech, and signal processing, ieee international conference on}. IEEE Computer Society, 1992, vol.~1, pp. 517--520.

\bibitem{charniak-johnson-2001-edit}
Eugene Charniak and Mark Johnson,
\newblock ``Edit detection and parsing for transcribed speech,''
\newblock in {\em NAACL}, 2001.

\bibitem{lee2020auxiliary}
Dongyub Lee, Byeongil Ko, Myeong~Cheol Shin, Taesun Whang, Daniel Lee, Eun~Hwa Kim, EungGyun Kim, and Jaechoon Jo,
\newblock ``Auxiliary sequence labeling tasks for disfluency detection,''
\newblock in {\em ICASSP 2020}.

\bibitem{ghosh2022span}
Sreyan Ghosh, Sonal Kumar, Yaman~Kumar Singla, Rajiv~Ratn Shah, and Srinivasan Umesh,
\newblock ``Span classification with structured information for disfluency detection in spoken utterances,''
\newblock {\em arXiv}, 2022.

\bibitem{DBLP:journals/corr/abs-1810-04805}
Jacob Devlin, Ming-Wei Chang, Kenton Lee, and Kristina Toutanova,
\newblock ``Bert: Pre-training of deep bidirectional transformers for language understanding,''
\newblock {\em arXiv preprint arXiv:1810.04805}, 2018.

\bibitem{DBLP:journals/corr/abs-1907-11692}
Yinhan Liu, Myle Ott, Naman Goyal, Jingfei Du, Mandar Joshi, Danqi Chen, Omer Levy, Mike Lewis, Luke Zettlemoyer, and Veselin Stoyanov,
\newblock ``Roberta: A robustly optimized bert pretraining approach,''
\newblock {\em arXiv preprint arXiv:1907.11692}, 2019.

\bibitem{clark2020electra}
Kevin Clark, Minh-Thang Luong, Quoc~V Le, and Christopher~D Manning,
\newblock ``Electra: Pre-training text encoders as discriminators rather than generators,''
\newblock {\em arXiv preprint arXiv:2003.10555}, 2020.

\end{thebibliography}

\end{document}